\documentclass{article}

\usepackage{hyperref}       
\usepackage{booktabs}       
\usepackage{amsfonts}       
\usepackage{nicefrac}       
\usepackage{microtype}      
\usepackage{xcolor}         
\usepackage{spconf,amsmath,graphicx}
\usepackage{graphicx}
\usepackage{multirow}
\usepackage{array}
\usepackage{amsmath}
\usepackage{amsfonts,amssymb}
\usepackage{bbding}

\title{Few-shot learning with improved local representations via bias rectify module}
\name{Chao Dong$^{1}$ \qquad Qi Ye$^{1}$ \qquad Wenchao Meng\thanks {$\star$ Corresponding author.}$^{1 \star}$ \qquad Kaixiang Yang$^{1}$}
\address{ $^{1}$Zhejiang University}

\begin{document}
%
\maketitle
\begin{abstract}
Recent approaches based on metric learning have achieved great progress in few-shot learning. However, most of them are limited to image-level representation manners, which fail to properly deal with the intra-class variations and spatial knowledge and thus produce undesirable performance. In this paper we propose a Deep Bias Rectify Network (DBRN) to fully exploit the spatial information that exists in the structure of the feature representations. We first employ a \emph{bias rectify module} to alleviate the adverse impact caused by the intra-class variations. \emph{bias rectify module} is able to focus on the features that are more discriminative for classification by given different weights. To make full use of the training data, we design a prototype augment mechanism that can make the prototypes generated from the support set to be more representative. To validate the effectiveness of our method, we conducted extensive experiments on various popular few-shot classification benchmarks and our methods can outperform state-of-the-art methods .
\end{abstract}

\section{Introduction} \label{intro} 

Few-shot learning (FSL) aims to learn a model with good generalization capability to get rid of the dependency of the annotated data. Concretely, it can be readily adapted to the agnostic tasks with a handful of labeled examples. However, the extremely limited annotated samples per class can hardly predict the truly class distribution, making FSL tasks challenging. 

To tackle the FSL problem, a variety of approaches have been proposed. For example, some state-of-the-art methods resort to learn a deep embedding network to represent the inter-class diversity\cite{VinyalsBLKW16,li2019DN4,SungYZXTH18}. Furthermore, they use non-parametric classifiers(e.g., the nearest neighbor classifier) to avoid the complex optimization problem in learning a classifier from a few examples. Another category of methods\cite{FinnAL17,antoniou2018how} constructs a meta-learner that can quick adopt to new tasks with a few labeled samples, either by a good initialization ,or by  effective learning algorithms. Moreover, the last group of methods\cite{HariharanG17,WangGHH18} directly solve the data deficiency by hallucinating new images based on labeled images from similar categories. 

The previous methods mainly focus on knowledge generalization, samples generating or optimization algorithms, but have not paid sufficient attention to the way of generating appropriate representations. Such representations can make better use of the limited examples. DN4\cite{li2019DN4} goes one step further, they replace the image-level representations with the set of local representations. This approach can preserve the discriminative knowledge which loses during the global pooling process. However, this method has not taken account of the spatial information, \emph{e.g.}, the background clutter and the intra-class variations. Such variations may force the features from the same class far away from each other in a given metric space. If there are sufficient labeled data for training, the subsequent learning procedure of convolutional neural networks with sufficient training samples can alleviate such interference. However, considering the specific nature of few-shot learning, it is nearly impossible to eliminate the impact caused by the noise and thus deteriorates the performance. Therefore, a desirable meta-learning approach should have the ability to properly utilize the spatial knowledge and reduce the interference caused by the aforementioned reasons. 

A simple approach to alleviate the noise is giving higher weight to the visual features are most discriminative for a given feature map. Towards this, we propose a deep bias rectify module based on non-local attention. The module will calculate the weight of each component by comparing it with the whole representations. Besides, the bags-of-features model used in DN4\cite{li2019DN4} takes all raw features generated from the support set for classification. Such features contain a lot of noise which affects the performance. Therefore, we assembled the feature embeddings as prototypes for each visual category similar to ProtoNet\cite{SnellSZ17}. In addition, we propose a simple approach called prototype augment to reduce the scarcity of data. We resize each support image to multi-scale before sending it to the feature extractor, the output feature maps will be fused to produce single prototype for classification.

\section{Our Method}

\begin{figure}
  \centering
  \includegraphics[width=0.5\textwidth]{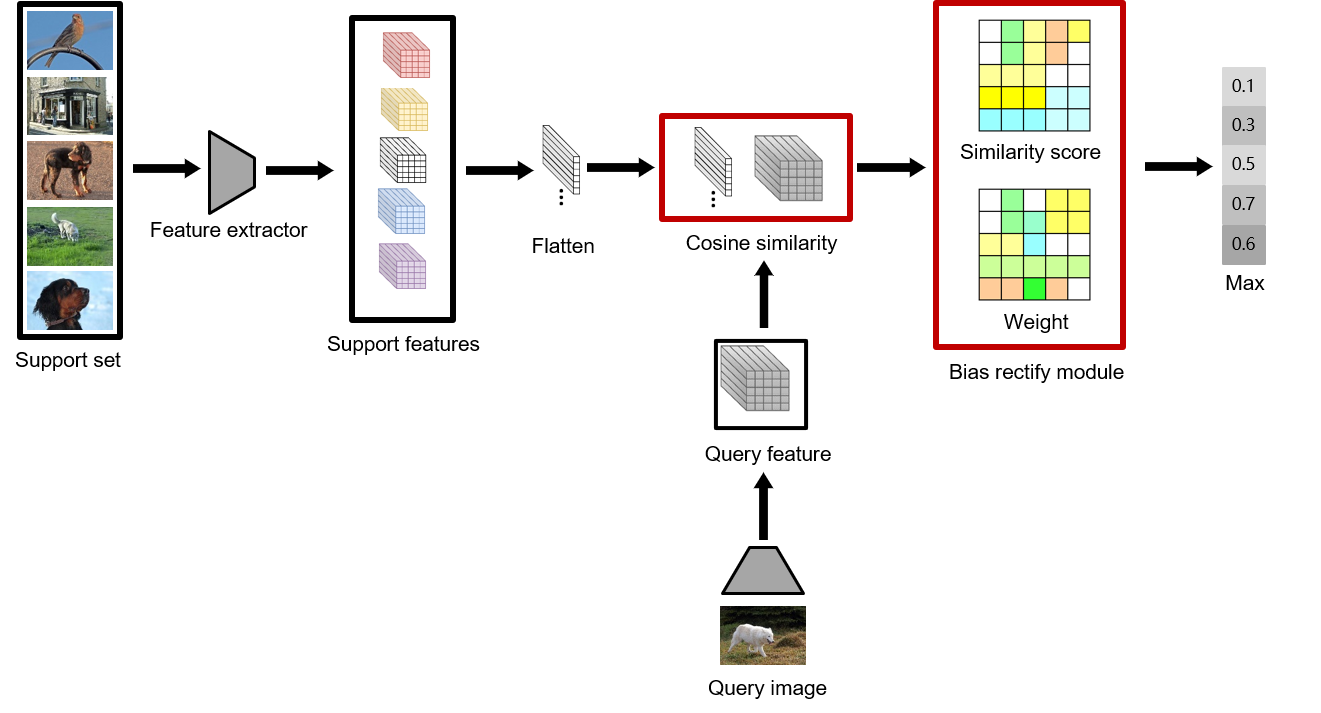}
  \caption{Illustration of the whole network architecture of our DBRN for a 5-way 1-shot few-shot classification task.}   
\label{fig:1}
	
\end{figure}

\subsection{Feature extractor} \label{Section:3.2.1}

As the traditional set in few-shot learning, we employ a CNN as the feature extractor which represents the images from support and query set by high dimensional feature. In general, the feature extractor only contains convolutional layers but has no fully connected layers $\Psi(\cdot)$. Unlike recent literature on the few-shot classification,  we remove the final global average pooling layer and produce a feature map $\Psi(x)\in \mathbb{R}^{w\times h\times d}$ rather than a vector $\Psi(x)\in \mathbb{R}^d$ to represent the visual feature of the image, where $w$ denotes the width and $h$ denotes the height of the feature map. Then the feature map of the image is formulated as:

\begin{equation}\label{equation:1}
 \Psi(x_i)\ =\ [v_1,...,v_r]\in \mathbb{R}^{r\times d},(r=w\times h)
\end{equation}

Where $u_i$ denotes the $i_{th}$ local feature of image and $r$ denotes the resolution of feature maps. Not in line with DN4\cite{li2019DN4} that directly use the local features for classification, we take the centriod of the local features as the \emph{prototype} of each class. We believe that take the centriod approach can eliminate the bias without extra parameters, which first proposed in ProtoNet\cite{SnellSZ17}. The difference is that we take the centriod of every local features. Based on Equation \ref{equation:1}, the \emph{prototype} of class $c$ in the support set can be formulated as:

\begin{equation}\label{equation:2}
 P_c\ =\ \frac{1}{|S_c|}\sum_{(x,y)\in S_c}{\Psi(x)}\in \mathbb{R}^{r\times d}
\end{equation}

Where $(x,y)\in S_c$ denotes the support examples  belong to class $c$, the $P_c$ can represent the visual feature of each class in the support set. The \emph{prototype} is used to calculate the similarity between images from the query set and support set.

\subsection{Similarity module}

The similarity module is used to calculate the distance between the prototype in Equation \ref{equation:2} and the query images to get the classification score of the query images. In this paper, we use the cosine similarity as metric function and \emph{k}-NN as similarity function to select the similarity value for query images.

As it described in Section \ref{Section:3.2.1}, given a query image $x$, the feature extractor will embed it as $\Psi(x)\ =\ [v_1,...,v_r]$. Then the similarity value can be calculated as follow, where The $P_c = [u_1,...,u_r]\in{\mathbb{R}^{r\times{d}}}$ means the \emph{prototype} for images belong to class $c$ and $k$ means we just concern the $k$ nearest feature vectors in the \emph{prototype} for the query image. The similarity between $x$ and $P_c$ is formulated as: 

\begin{equation}  \label{equation:3}
	\begin{split}
 	& Sim\left(P_c,\Psi\left(x\right)\right)=\sum_{j=0}^{k} s_{\tau}\left(u_i,v_j\right)\\
	& s_{\tau}(x,y)=\tau \left<\hat{x},\hat{y}\right>
	\end{split}
\end{equation}

for $x,y\in{\mathbb{R}^d}$, $\hat{x}$ is the $l_2$-normalized conuterpart for $x$. $s_{\tau}$ is the scaled cosine similarity, with $\tau$ being a learnable parameter as in \cite{QiBL18,GidarisK18}.  Then, the similarity module is defined as:

\begin{equation} \label{equation:6}
	f_{\theta}[P](x)\ =\ \sigma\left([Sim\left(P_c,\Psi\left(x\right)\right)]_{c=1}^{n}\right)
\end{equation} 

Where $ P=[P_1,...,P_n] $ means the \emph{prototype set} for all classes in support set and $\sigma:\mathbb{R}^d\rightarrow\mathbb{R}^d$ is the \emph{softmax function}.

\begin{figure}
  \centering
  \includegraphics[width=0.4\textwidth]{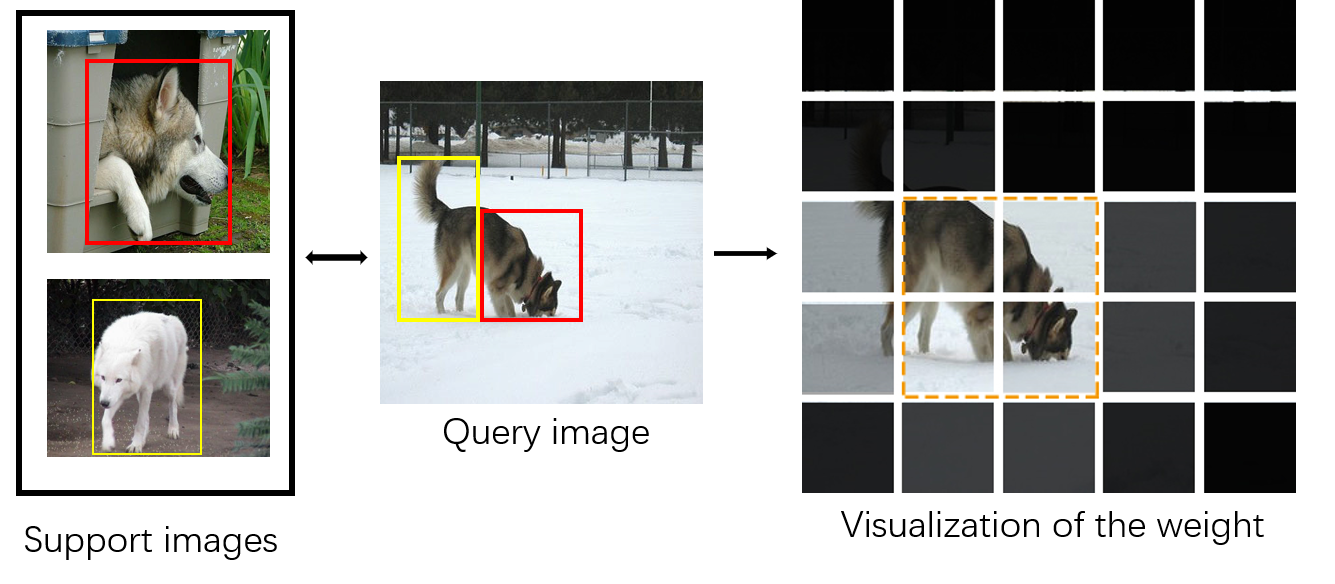}
  \caption{The illustration for the motivation for bias rectify module.}   
\label{fig:2}
	
\end{figure}

\subsection{Bias rectify module}  \label{Section:3.2.3}

In traditional few-shot classification tasks, the support set merely consists of several images for each class. Then we need to classify different query images based on this support set. The lack of data in the support set leads to an inevitable problem: the bias in different images from the support set and the query set will deteriorate the performance of the classification model.  As  Figure \ref{fig:2} shows, different parts of the object may exist in the support images and query images. It will lead to a significant difference in the visual features. So that we propose a bias rectify module that can effectively alleviate the noise and enhance the accuracy of our model. This is the most improvement we have made based on DN4\cite{li2019DN4}. 

Firstly, we can observe that the features which occupy a larger area in original images will occupy more space in feature maps. Therefore, we can calculate the co-occurrence rate of the special part images by calculating their corresponding feature map co-occurrence rate. To achieve this goal, we calculate the cosine similarity between the local feature of query images and the whole feature map to generate the co-occurrence rate of this local feature: 

\begin{equation}  \label{equation:4}
{W}_{v_j}\ =\frac{1}{\xi}\sum_{i\ =0}^{r}\left(\frac{u_i\times v_j}{\left|u_i\right|\times\left|v_j\right|}\right)^\omega
\end{equation}

Where the ${W}_{v_j}$ denotes the co-occurrence rate of local feature vector $v_i$ in query images, $r$ denotes the resolution of the feature maps generated from support images, $\xi$ denotes the normalized parameter and $\omega$ denotes a hyper-parameter that control the dispersion of ${W}_{v_j}$.

This bias rectify module is simple but effective during few-shot classification tasks that can benefit the experiment result without any extra parameters. This module will more effective when the support images and query images have more a different appearance.

\subsection{Prototype augment}

Consider the lack of labelled data, the size of the object in the support images may differ from the object in the query images. Therefore, we resize the support images in multi-scale before input them to the feature extractor, then we calculate the prototypes of the support set with all these feature maps. By this method, the prototypes of the support set become more robust when encountering the variation of the size of the object. The prototype generated from the support set can be represented as follow:

\begin{equation}   \label{equation:5}
Prototype(c)=\frac{1}{|S_c|}\sum_{P_{c_i}\in{S_c}}{P_{c_i}}
\end{equation}

Where $S_c=[P_{c_i},...,P_{c_n}]$ denote the set of prototypes with different scale, $P_{c_i}$ denote the \emph{prototype} of class $c$ with the $i_th$ scale which calculated by Equation \ref{equation:2}. In our experiments, we take the triple-scale method, which means the images in the support and query set will become three times during meta-training and testing. During experiment, we set the resolution of the images as $84\times84$, $92\times92$, $108\times108$ .

\section{Experiments}
\subsection{Dataset}
We use the ResNet-12 trained following previous work FEAT\cite{YeHZS20} on a RTX3090. 
We evaluate the performance of our method for few-shot classification tasks on three popular benchmark dataset: miniImageNet\cite{VinyalsBLKW16}, tieredImageNet\cite{RenTRSSTLZ18}, Caltech-USCD birds-200-2011(CUB)\cite{2011The}.

\emph{mini}ImageNet is derived from ILSVRC-12 dataset. It contains 100 different classes with 600 samples per class. We follow the splits used in previous  work \cite{RaviL17}, which divide the dataset into 64, 16, 20 for training, validation and test, respectively.
\emph{tiered}ImageNet is a larger subset derived from ILSVRC-12 dataset\cite{RussakovskyDSKS15}. It contains 608 classes from 34 super-classes, which includes 1281 images each class. We follow the splits in \cite{RenTRSSTLZ18}, where take 351, 97 and 160 classes as the training set, validation set and testing set respectively.
 CUB was originally proposed for fine-grained classification tasks. It includes 200 different birds with 11,788 images in total.  Following the splits in previous works \cite{ChenLKWH19}, we take 100 classes for training, 50 classes for validation and 50 classes for testing. 

\subsection{Implementation details}

 Because the motivation of this work is largely inspired by the DN4 \cite{li2019DN4}.  We re-implement the DN4 as our baseline model. We use the ResNet12\cite{GidarisK18} as our model backbone following the previous literature to get a fair comparison with previous work. For each dataset, we pre-train a classifier with the training set. Then we remove the fully connected layer in ResNet12, so that the network becomes a convolutional network that maps each input image as a feature vector. In order to preserve the local feature for input images, we remove the global average pooling layer so that the network becomes a fully convolutional network. When input images resize as $84\times 84$, the backbone network generates a feature map with size $5\times 5\times 640$.Different from previous works which training the network from scratch, we apply a pre-train strategy as suggested in \cite{QiaoLSY18}. After pre-training, we remove the fully connected layer select the pretrained model with the highest performance in validation set.  
 
\begin{figure}
  \centering
  \includegraphics[width=0.4\textwidth]{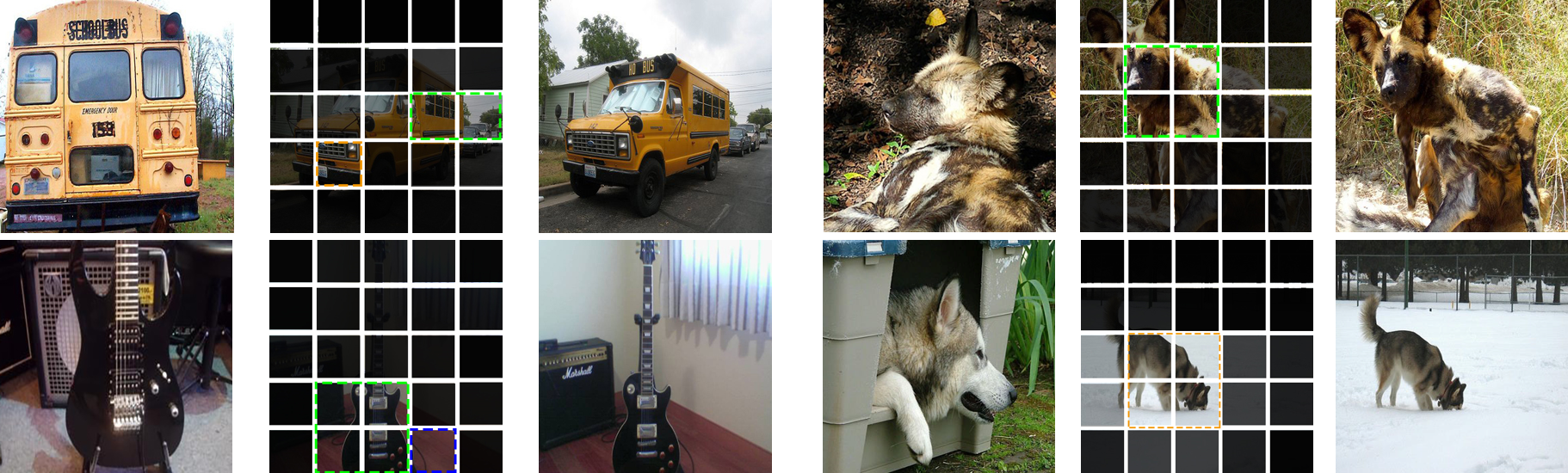}
  \caption{Visualization for our DBRN model.\textbf{Please zoom for details}.}   
\label{fig:4}
\end{figure}

\begin{table}\scriptsize
  \caption{Average 5-way classification performance($\%$) with 95$\%$ confidence intervals on \emph{mini}ImageNet and \emph{tiered}ImageNet.}
  \label{Table:2}
  \centering
  \setlength{\tabcolsep}{1mm}{
  \begin{tabular}{m{1.5cm} l m{1cm}<{\centering}m{1cm}<{\centering}m{1cm}<{\centering}m{1cm}<{\centering}}
  \toprule
  \multirow{2}{*}{Method}  &\multirow{2}{*}{backbone}   &\multicolumn{2}{c}{\textbf{\emph{mini}ImageNet}}      & \multicolumn{2}{c}{\textbf{\emph{tiered}ImageNet}} \\
  \cmidrule(r){3-6}
   && 1-shot  & 5-shot  &1-shot  &5-shot\\
  \midrule
	ProtoNet\cite{SnellSZ17}            &\emph {ResNet12}   &$60.37 \pm 0.83$       &$78.02 \pm 0.57$   
   &   $65.65 \pm 0.92 $      & $83.40 \pm 0.65$\\
   Baseline++\cite{ChenLKWH19}          &\emph {ResNet12}   & $55.43 \pm 0.81$   &$77.18\pm0.61$    
&$61.49 \pm 0.91$     &$82.37 \pm 0.67$\\
  Neg-Cosine\cite{LiuCLL0LH20}          &\emph {ResNet12}   &$63.85 \pm 0.81$   &$81.57 \pm 0.56$ & -  & -\\
	FEAT\cite{YeHZS20}            &\emph {ResNet12}   &$66.78 \pm 0.20$   &$82.05 \pm 0.15$ 
&$70.80 \pm 0.23$  &$84.79 \pm 0.16$\\
	CTM\cite{LiEDZW19}           &\emph {ResNet12}   &$64.12 \pm 0.82$   &$80.51 \pm 0.13$    
&$68.41 \pm 0.39$  &$84.28 \pm 1.73$\\
	DeepEMD\cite{ZhangCLS20}           &\emph {ResNet12}   &$65.91 \pm 0.82$   &$82.41\pm0.56$    
&$71.16\pm0.87$  &$86.03 \pm 0.58$\\
	E3BM\cite{LiuSS20}            &\emph {ResNet12}   &$63.80 \pm 0.40$   &$80.10 \pm 0.30$ 
&$71.20 \pm 0.40$  &$85.30 \pm 0.30$\\
	PPA\cite{QiaoLSY18}            &\emph {WRN-28-10}   &$59.60\pm -$  &$77.46 \pm - $
&$63.99 \pm -$  &$81.97 \pm -$\\
	LEO\cite{rusu2018metalearning}            &\emph {WRN-28-10}	&$61.76 \pm 0.08$	  &$77.59 \pm 0.12$	
&$66.33 \pm 0.05$    &$81.44 \pm 0.09$\\
	TADAM\cite{OreshkinLL18}	       &\emph {ResNet12}	&$58.50 \pm 0.30$   &$76.70 \pm 0.30$	&-	&-\\
  \midrule
   \textbf {Our Baseline}	&\emph{ResNet12}	&$61.13 \pm 0.27$	 &$76.97 \pm 0.20$	
&$70.45 \pm 0.22$	 &$83.37 \pm 0.33$\\
	\textbf{DBRN(Ours)}	    &\emph{ResNet12}     &$\textbf{67.01} \pm \textbf{0.28}$	&$\textbf{83.33} \pm \textbf{0.19}$	   &$\textbf{72.80} \pm \textbf{0.31}$	&$\textbf{87.13} \pm \textbf{0.21}$\\
  \bottomrule
 \end{tabular}}
\end{table}

\begin{table}\scriptsize
  \caption{Average 5-way classification performance($\%$) with 95$\%$ confidence intervals on CUB.}
  \label{Table:3}
  \centering
  \setlength{\tabcolsep}{1mm}{
  \begin{tabular}{m{2cm} m{1.5cm} m{2cm}<{\centering}m{2cm}<{\centering}}
  \toprule
  \multirow{2}{*}{Method}  &\multirow{2}{*}{backbone}   &\multicolumn{2}{c}{\textbf{CUB}}       \\
  \cmidrule(r){3-4}
   && 1-shot  & 5-shot  \\
  \midrule
	ProtoNet\cite{SnellSZ17}            &\emph {ResNet12}   &$66.09\pm0.92$       &$82.50\pm0.58$   \\
   Baseline++\cite{ChenLKWH19}          &\emph {ResNet12}   & $67.02\pm0.90$   &$83.58\pm0.54$    \\
  Neg-Cosine\cite{LiuCLL0LH20}          &\emph {ResNet12}   &$72.66\pm0.85$   &$89.40\pm0.43$ \\
	MAML\cite{FinnAL17}            &\emph {Conv4}   &$50.45\pm0.97$   &$59.60\pm0.84$ \\
	MVT\cite{ParkHBKSLHH20}           &\emph {ResNet12}   &-   &$80.33 \pm 0.61$    \\
	DeepEMD\cite{ZhangCLS20}            &\emph {ResNet12}   &$75.65\pm0.83$   &$88.69\pm0.50$    \\
  \midrule
   \textbf {Our Baseline}	&\emph{ResNet12}	&$66.63\pm0.28$	 &$81.64\pm0.19$	\\
	\textbf{DBRN(Ours)}	    &\emph{ResNet12}     &$\textbf{75.78}\pm \textbf{0.27}$	
&$\textbf{92.21}\pm \textbf{0.14}$	  \\
  \bottomrule
 \end{tabular}}
\end{table}

\begin{table}\scriptsize
\caption{Ablation study on miniImageNet with 95\% confidence intervals.}
\label{Table:5}
  \centering
  \setlength{\tabcolsep}{1mm}{
  \begin{tabular}{m{1.5cm}<{\centering} m{1.5cm}<{\centering} m{1.5cm}<{\centering}m{1.5cm}<{\centering} m{1.5cm}<{\centering} }
  \toprule
  \multirow{2}{*}{Pow} &\multirow{2}{*}{Weight} &\multirow{2}{*}{ProtoAug} &\multicolumn{2}{c}{\emph{mini}ImageNet}\\
  \cmidrule(r){4-5}
  &&&5way1shot  &5way5shot  \\
  \midrule
  \XSolidBrush &\XSolidBrush &\XSolidBrush &63.88 &80.52\\
  \XSolidBrush &\Checkmark &\XSolidBrush &65.53 &81.80\\
  \Checkmark &\Checkmark &\XSolidBrush &66.03 &82.58\\
  \Checkmark &\Checkmark &\Checkmark &67.01 &83.33\\
  \bottomrule
  \end{tabular}}
\end{table}

\subsection{Main results and Comparisons}  \label{Section:4.2}

Table \ref{Table:2} and Table \ref{Table:3} presents 5-way classification accuracy ($\%$) with 95$\%$ confidence intervals of our method and others on miniImageNet, tieredImageNet and CUB. We take the \emph{k}-NN algorithm similar to DN4\cite{li2019DN4} as our baseline, which selects the k nearest feature vectors between support set and query image as similarity vector. Firstly, we can observe that our baseline already archives a better performance than some state-of-the-art methods. This may be because we take an end-to-end training strategy rather than fix the backbone network and temperature scaling of the logits while meta-training. Moreover, our DBRN further promotes the performance and outperforms all state-of-the-art methods with a significant margin which effectively demonstrate the effectiveness of our method. Compared our DBRN with our baseline, we can observe that apply an original k-NN strategy on backbone for few-shot classification is not enough. It means that simply represent similarity for two images with nearest features discard important knowledge between images. 

\subsection{Ablative study}   \label{Section:4.3}

To further validate the effectiveness of our method, we apply ablation experiments as it is shown in Table \ref{Table:5}. Weight and pow are components for the bias rectify module in Equation \ref{equation:4} . Protoaug is prototype augment mechanism in Equation \ref{equation:5}.  The ablation of each component in our model will result in the drop of performance.

\subsection{Visualization of bias rectify weights}

To demonstrate the effectiveness of the bias rectify module proposed in Section \ref{Section:3.2.3} during the inference process, we conduct a visualization experiment between regions of correspondence images for support set and query set. We give different brightness for the regions of query images concerning their weight.  As it is shown in Figure \ref{fig:4} , the regions which co-occurrence in both images will get higher weight,  regardless of the size difference between these regions. That means bias rectify module will alleviate the bias between different images and benefit the classification performance. We take both correctly and incorrectly classified pair of images in our experiment to further demonstrate the robustness of our method. 

\section{Conclusion and future work}

In this work, we proposed a simple but effective DBRN for the few-shot classification. We used feature extractors based on end-to-end training on base-class with standard cross-entropy loss without any extra data or complex meta-training strategy that advocated in recent few-shot literature. Firstly, we focus on the intra-class variations and high-variance background that exists in the support and query set and proposed a nonparametric rectify module to alleviate the influence of this bias. Then prototype augment mechanism is proposed for the lack of the labeled data during inference. In this context, designing a more powerful prototype generation mechanism that can effectively utilize the discriminative information between images looks like a very promising direction for future research. The experiment results obtained on three popular datasets demonstrated that our DBRN  significantly outperforms existing methods and achieves new state-of-the-art on few-shot classification task.

\bibliographystyle{IEEEbib}
\bibliography{refs}

\end{document}